\documentclass[conference]{IEEEtran}
\usepackage[T1]{fontenc}
\usepackage{lmodern}
\usepackage{amsmath,amssymb}
\usepackage{graphicx}
\usepackage{booktabs}
\usepackage{multirow}
\usepackage{xcolor}
\usepackage{microtype}
\usepackage{siunitx}
\usepackage{subcaption}
\usepackage{threeparttable}
\usepackage[hidelinks]{hyperref}
\hypersetup{pdftitle={Asking a Multimodal LLM Whether Someone Fell: Zero-Shot Screening of Low-Resolution Thermal Video in a Laboratory and an Occupied Home},
  pdfauthor={Marcos Junca, Mauro Junca},
  pdfsubject={Zero-shot fall screening of low-resolution thermal video with a multimodal large language model (Gemini), evaluated on a public thermal dataset and in an occupied home},
  pdfkeywords={fall detection, thermal imaging, thermopile array, multimodal large language model, vision-language model, Gemini, ambient assisted living, privacy-preserving sensing, false alarms}}
\usepackage[capitalise,noabbrev]{cleveref}
\graphicspath{{figs/}}
\newcommand{\BedBase}{0/24}

\newcommand{\ClipsPerHour}{9.8}
\newcommand{\CostHome}{0.0091}
\newcommand{\CostLab}{0.0044}

\newcommand{\EhomeCoreRecovery}{315}
\newcommand{\EhomeCoreRest}{23}
\newcommand{\EhomeCoreUpright}{161}
\newcommand{\EhomeDetectorAgree}{73}
\newcommand{\EhomeFallWindows}{438}
\newcommand{\EhomeFpsMedian}{15.5}
\newcommand{\EhomeFrames}{92908}
\newcommand{\EhomeNegWindows}{563}
\newcommand{\EhomeRecovery}{335}
\newcommand{\EhomeRest}{46}
\newcommand{\EhomeUpright}{182}
\newcommand{\EitherHomeCalls}{0}
\newcommand{\EitherHomeClips}{73}
\newcommand{\EitherHomeFirstCalls}{0}
\newcommand{\EitherHomeFirstVisible}{4/9}
\newcommand{\EitherHomeVisible}{5/9}
\newcommand{\EitherLabFA}{0}
\newcommand{\EitherLabFirstSens}{67}
\newcommand{\EitherLabN}{499}
\newcommand{\EitherLabSens}{72}

\newcommand{\FABaseHeldout}{0.0}

\newcommand{\FAFlashThreeFiveHeldout}{0.0}

\newcommand{\FARuleHeldout}{1.1}

\newcommand{\FATreesHeldout}{1.1}

\newcommand{\FAcountBaseFull}{12}

\newcommand{\FAnBaseAll}{499}

\newcommand{\GateLie}{2/2}
\newcommand{\GateOccluded}{15/17}
\newcommand{\GotUpCount}{170/273}
\newcommand{\GotUpSens}{62}
\newcommand{\GroundingBase}{99}
\newcommand{\GroundingCount}{276/279}
\newcommand{\HomeAftermathClips}{22}
\newcommand{\HomeBaseAftermath}{0/22}

\newcommand{\HomeBaseCallsWord}{none}
\newcommand{\HomeBaseClips}{395}
\newcommand{\HomeBaseConfirmedTestCalls}{1}
\newcommand{\HomeBaseConfirmedTestClips}{103}
\newcommand{\HomeBaseLie}{0/2}
\newcommand{\HomeBaseOccluded}{0/15}

\newcommand{\HomeBasePerDayHi}{1.2}

\newcommand{\HomeBaseSampleCalls}{0}

\newcommand{\HomeBaseVisible}{4/9}
\newcommand{\HomeBaseVisibleK}{4}
\newcommand{\HomeCameraHours}{98.2}
\newcommand{\HomeClips}{958}
\newcommand{\HomeConfirmedTestHours}{4.0}
\newcommand{\HomeFfcPerHour}{28}
\newcommand{\HomeFirstDay}{6 September 2026}
\newcommand{\HomeFpsMedian}{8.8}
\newcommand{\HomeLastDay}{13 September 2026}
\newcommand{\HomeLieClips}{2}

\newcommand{\HomeNoAudioVisible}{2/9}

\newcommand{\HomeNoOovCallsWord}{none}
\newcommand{\HomeNoOovClips}{73}
\newcommand{\HomeNoOovLie}{2/2}
\newcommand{\HomeNoOovOccluded}{13/15}

\newcommand{\HomeNoOovVisible}{9/9}
\newcommand{\HomeNoOovVisibleK}{9}
\newcommand{\HomeNormalClips}{395}
\newcommand{\HomeNormalHours}{76}
\newcommand{\HomeNormalHoursWithConfirmed}{80}
\newcommand{\HomeOccludedClips}{15}
\newcommand{\HomeOccupiedPct}{10}
\newcommand{\HomeRecordedHours}{98.4}

\newcommand{\HomeResEightBySixCalls}{0}

\newcommand{\HomeResEightBySixVisible}{0/9}

\newcommand{\HomeRuleCalls}{76}

\newcommand{\HomeRulePerDay}{23.9}

\newcommand{\HomeRuleVisible}{7/9}

\newcommand{\HomeSegments}{5906}

\newcommand{\HomeSevBaseCalls}{0}
\newcommand{\HomeSevBaseCallsWord}{none}
\newcommand{\HomeSevBaseClips}{73}

\newcommand{\HomeSevBaseOccluded}{2/15}

\newcommand{\HomeSevBaseVisible}{6/9}

\newcommand{\HomeSevNoAudioVisible}{5/9}

\newcommand{\HomeSevNoOovCalls}{1}

\newcommand{\HomeSevNoOovLie}{0/2}
\newcommand{\HomeSevNoOovOccluded}{11/15}

\newcommand{\HomeSevNoOovVisible}{9/9}

\newcommand{\HomeSevResEightBySixCalls}{0}

\newcommand{\HomeSevResEightBySixVisible}{3/9}

\newcommand{\HomeSevResThirtyTwoByTwentyFourCalls}{1}

\newcommand{\HomeSevResThirtyTwoByTwentyFourVisible}{7/9}

\newcommand{\HomeSevVfiveCalls}{0}

\newcommand{\HomeSevVfiveLie}{1/2}

\newcommand{\HomeSevVfiveNoAudioOccluded}{8/15}

\newcommand{\HomeSevVfiveOccluded}{11/15}

\newcommand{\HomeSevVfiveVisible}{6/9}

\newcommand{\HomeTestHours}{10}
\newcommand{\HomeTestSessions}{8}

\newcommand{\HomeTreesCalls}{45}

\newcommand{\HomeTreesClips}{395}

\newcommand{\HomeTreesPerDay}{14.2}

\newcommand{\HomeTreesVisible}{6/9}

\newcommand{\HomeVfiveCalls}{0}

\newcommand{\HomeVfiveLie}{2/2}

\newcommand{\HomeVfiveNoAudioLie}{0/2}
\newcommand{\HomeVfiveNoAudioOccluded}{5/15}

\newcommand{\HomeVfiveOccluded}{10/15}

\newcommand{\HomeVfiveVisible}{5/9}

\newcommand{\HomeVisibleClips}{9}

\newcommand{\LatencyHome}{21.3}
\newcommand{\LatencyLab}{9.2}
\newcommand{\LatencyNinetyHome}{74.1}
\newcommand{\LatencyNinetyLab}{24.8}
\newcommand{\MissLowering}{87}

\newcommand{\MissOutOfView}{34}
\newcommand{\MissTotal}{135}

\newcommand{\PairGraySevenHoSensA}{80}
\newcommand{\PairGraySevenHoSensB}{73}

\newcommand{\PairModelN}{414}

\newcommand{\PairModelP}{<0.001}
\newcommand{\PairModelSensA}{67}
\newcommand{\PairModelSensB}{77}
\newcommand{\PairOovEightHoFaA}{0}
\newcommand{\PairOovEightHoFaB}{0}
\newcommand{\PairOovEightHoN}{140}
\newcommand{\PairOovEightHoNegN}{182}

\newcommand{\PairOovEightHoP}{<0.001}
\newcommand{\PairOovEightHoSensA}{69}
\newcommand{\PairOovEightHoSensB}{83}
\newcommand{\PairOovSevenFaA}{0}
\newcommand{\PairOovSevenFaB}{4}

\newcommand{\PairOovSevenHoSensA}{80}
\newcommand{\PairOovSevenHoSensB}{92}
\newcommand{\PairOovSevenN}{414}
\newcommand{\PairOovSevenNegN}{499}
\newcommand{\PairOovSevenOovA}{24}
\newcommand{\PairOovSevenOovB}{0}
\newcommand{\PairOovSevenP}{<0.001}
\newcommand{\PairOovSevenSensA}{77}
\newcommand{\PairOovSevenSensB}{88}

\newcommand{\PairResEightEightHoFaB}{2}

\newcommand{\PairResEightEightHoNegN}{182}

\newcommand{\PairResEightEightHoSensA}{69}
\newcommand{\PairResEightEightHoSensB}{34}

\newcommand{\PairResEightEightOovA}{35}
\newcommand{\PairResEightEightOovB}{132}

\newcommand{\PairResEightSevenHoFaB}{5}

\newcommand{\PairResEightSevenHoSensB}{51}

\newcommand{\PairResSixteenEightHoSensB}{48}

\newcommand{\PairResSixteenSevenHoSensA}{80}
\newcommand{\PairResSixteenSevenHoSensB}{61}

\newcommand{\PairVfiveEightHoFaB}{2}
\newcommand{\PairVfiveEightHoN}{140}
\newcommand{\PairVfiveEightHoNegN}{182}

\newcommand{\PairVfiveEightHoP}{=0.007}
\newcommand{\PairVfiveEightHoSensA}{69}
\newcommand{\PairVfiveEightHoSensB}{79}

\newcommand{\PairVfiveSevenHoFaB}{0}

\newcommand{\PairVfiveSevenHoSensA}{80}
\newcommand{\PairVfiveSevenHoSensB}{81}

\newcommand{\RetestLabAgree}{95.9}
\newcommand{\RetestLabFlips}{41}
\newcommand{\RetestLabFlipsPct}{4}
\newcommand{\RetestLabKappa}{0.90}
\newcommand{\RetestLabN}{1{,}001}
\newcommand{\RetestSevLabAgree}{95.8}

\newcommand{\RetestSevLabKappa}{0.91}
\newcommand{\RetestSevLabN}{360}
\newcommand{\SensBaseAll}{67}

\newcommand{\SensBaseHeldout}{69}

\newcommand{\SensClusterBaseAll}{62--73}

\newcommand{\SensCountBaseAll}{279/414}
\newcommand{\SensCountBaseHeldout}{97/140}

\newcommand{\SensFlashThreeFiveHeldout}{86}

\newcommand{\SensRuleHeldout}{94}

\newcommand{\SensSevBaseHeldout}{80}

\newcommand{\SensSevDisplayHeldout}{86}

\newcommand{\SensTreesHeldout}{100}

\newcommand{\StillDownCount}{109/141}
\newcommand{\StillDownSens}{77}
\newcommand{\StudyCost}{65}
\newcommand{\StudyTokens}{70}
\newcommand{\SubjectSensMax}{83}
\newcommand{\SubjectSensMin}{49}
\newcommand{\TokensHome}{10,909}
\newcommand{\TokensLab}{4,394}

\newcommand{\TotalInvalid}{0}

\newcommand{\VetoHomeDescribed}{16}
\newcommand{\VetoHomeMissed}{20}
\newcommand{\VetoHomeOutOfView}{17}
\newcommand{\VetoLabDescribed}{28}

\newcommand{\WakesBudget}{65}
\newcommand{\WakesOpened}{543}
\newcommand{\WakesPerHour}{38.4}
\newcommand{\WakesSoundPct}{98}
\newcommand{\WakesTotal}{3,769}

\begin{document}

\title{Asking a Multimodal LLM Whether Someone Fell:\\Zero-Shot Screening of Low-Resolution Thermal\\Video in a Laboratory and an Occupied Home}

\author{\IEEEauthorblockN{Marcos Junca and Mauro Junca}
\IEEEauthorblockA{Senecta Inc.\\ marcos@senecta.ai}}

\maketitle

\begin{abstract}
Thermal cameras can watch over older people living alone without capturing faces, and multimodal large
language models (LLMs) can judge a short video without task-specific training. We evaluate a production
multimodal LLM, Gemini~3.8 Flash at temperature 0, as the decision stage of an event-triggered fall
screener, used exactly as a deployed prototype uses it: one request per short thermal clip, with its sound
track when there is one, and its predecessor Gemini~3.7 Flash on the same requests. On \EhomeFallWindows{}
staged falls and \EhomeNegWindows{} non-fall windows that we located by hand in a public $32\times24$ thermal
dataset, the deployed request called \SensBaseHeldout\% of the held-out subjects' falls (\SensBaseAll\% over
all subjects; bed falls excluded) and none of \FAnBaseAll{} non-fall windows, including people lying on the
floor and getting up. Over \HomeNormalHours{}\,h of ordinary life in an occupied room, recorded at
$160\times120$ with sound, it called \HomeBaseCallsWord{} of \HomeBaseClips{} clips a fall (95\% upper
bound \HomeBasePerDayHi{} false alarms per day) and caught \HomeBaseVisibleK{} of the
\HomeVisibleClips{} staged falls the camera could see. One cautionary sentence in the prompt (``a person
who goes out of view \ldots\ is not a fall'') accounted for many of the misses: removing it raised
sensitivity on the held-out subjects from \PairOovEightHoSensA\% to \PairOovEightHoSensB\% on 3.8 Flash and
from \PairOovSevenHoSensA\% to \PairOovSevenHoSensB\% on Gemini~3.7 Flash, and in the home from
\HomeBaseVisibleK{} to \HomeNoOovVisibleK{} of the \HomeVisibleClips{} visible falls, with no false call
in ordinary living but at the price of calling deliberate lie-downs a fall. Sensitivity collapsed at $8\times6$ pixels
(\PairResEightEightHoSensB\%); identical requests disagreed on \RetestLabFlipsPct\% of windows; and when the
model called a fall, its
explanation placed it within a second of the annotated landing \GroundingBase\% of the time. Two
conventional detectors trained on the laboratory data caught nearly every held-out laboratory fall but
raised \HomeTreesPerDay{} and \HomeRulePerDay{} false alarms per day in the home. We release the
frame-level fall annotations of the public dataset and every model answer on it.
\end{abstract}


\section{Introduction}
\label{sec:intro}

Falls are a leading cause of injury among older adults, and what happens after the fall matters as
much as the fall itself. In a cohort of people over 90, about 80\% of those who fell could not get up
unaided after at least one fall, three in ten of them lay on the floor for an hour or more, over 80\% of
falls happened when they were alone, and call alarms, though widely available, went unused in most
falls~\cite{fleming2008inability}. A system that notices a fall and asks the person whether they are all
right could shorten that time. Wearable alarms only help when they are worn and pressed, and cameras are
unwelcome in many homes. Thermal sensing sits in between: a low-resolution thermal image shows a warm body and its
posture, but not a face or a document~\cite{sixsmith2004smart}.

Thermal fall detection has been studied for two decades, almost always by training a classifier on
staged falls and testing it on more staged falls from the same setting~\cite{mashiyama2014fall,
vadivelu2017thermal,nogas2020deepfall,riquelme2019ehomeseniors}. Two problems recur. Staged falls by
young adults differ from real falls of older people~\cite{bagala2012evaluation,klenk2011comparison},
and in a real home the decisive number is how often the system raises an alarm when nothing happened,
which laboratory datasets cannot measure. A screener that interrupts a household twice a day is switched
off.

Multimodal large language models (LLMs) offer a different route: they can watch a short video with its
sound track and answer a question about it without any task-specific training, and they explain their
answer in words. This makes them attractive as the \emph{decision stage} of a screening pipeline. A cheap
always-on gate decides that something happened, and the model is asked once about a short clip. But
thermal video is far from the visible-light video these models were trained on, and it is not obvious
how they behave on a $32\times24$ thermopile array or a $160\times120$ microbolometer, how much the
answer depends on how the temperatures are rendered into colours, on a sentence of the prompt, or on the
sound, and how stable the answer is when the same question is asked twice.

We measure these things for a production model, Gemini~3.8 Flash, used exactly as a deployed prototype
uses it. Our contributions are:
\begin{itemize}
\item An evaluation of the LLM review stage on two complementary sources: \EhomeFallWindows{}
hand-located staged falls and \EhomeNegWindows{} non-fall windows from a public $32\times24$ thermal
dataset~\cite{riquelme2019ehomeseniors}, with held-out subjects; and every clip that an event-triggered
pipeline produced over \HomeCameraHours{} camera-hours of continuous radiometric thermal video and audio
in one occupied room.
\item Single-factor ablations of sensor resolution (from $160\times120$ down to $8\times6$), colour map,
display size, sound, prompt wording and model generation, plus repeated identical requests to measure
test--retest stability at temperature 0.
\item A comparison with conventional detectors (a hand-written descent rule and a gradient-boosted
classifier trained on the public data), in the laboratory and transferred to the home.
\item Practical lessons for anyone using an LLM as a sensor's judge, including how one cautious sentence
in the prompt, which kept false alarms at home near zero, also cost a large share of the visible
laboratory falls.
\end{itemize}
We release the frame-level fall annotations of the public dataset, the windows cut from it, every
prompt and every model answer on those windows, under CC~BY~4.0, at
\url{https://senecta.ai/research/thermal-fall-llm-data.zip}. The home recordings stay private.

\section{Related work}
\label{sec:related}

\paragraph{Thermal and thermopile fall detection}
Infrared sensing for falls goes back to SIMBAD, a low-cost pyroelectric array with a field trial in
homes~\cite{sixsmith2004smart}. Most later work trains a classifier on staged falls: hand-crafted features
on a low-resolution array~\cite{mashiyama2014fall}, optical flow on a handheld thermal camera and the
Thermal Simulated Fall dataset~\cite{vadivelu2017thermal}, autoencoders trained on normal activity that
treat a fall as an anomaly~\cite{nogas2020deepfall,mehta2021motion}, recurrent and 3D convolutional
networks on very-low-resolution arrays~\cite{taramasco2018novel,naser2023privacy}, including $32\times24$
arrays like the one in our laboratory data~\cite{tateno2020privacy,rezaei2023unobtrusive}, and recent
large datasets and models: TF-66 with 66 participants in nine environments~\cite{silver2025tf66,
silver2025biconvlstm} and TaFall, an $80\times62$ array evaluated on more than 3{,}000 falls and in four
homes~\cite{li2026tafall}. MUVIM recorded falls with thermal, depth, infrared and RGB cameras
together~\cite{denkovski2022multi}. The eHomeSeniors dataset~\cite{riquelme2019ehomeseniors} provides
the staged falls we use; the one follow-up we found that reports results on it trains a support vector
machine on its frames~\cite{miawarni2022enhancing}. Supervised models reach high accuracy on such data;
the cost is training data from the target sensor, and the open question is what happens elsewhere.
Two systems close to ours add a second stage after the thermal classifier: a human confirmation step to
cut false positives~\cite{naser2023privacy}, and a voice assistant with an LLM that talks to the person
after the fall~\cite{lupion2025privacy}; in both, the language model never sees the thermal video.
Low resolution is also argued for on privacy grounds: recognition survives at $16\times12$
pixels~\cite{ryoo2017privacy}, CNN activity recognition degrades as a thermal array is reduced from
$32\times32$ to $4\times4$~\cite{anyanwu2022thermal}, and older adults raise few privacy concerns about
silhouette-like images~\cite{demiris2009older}.

\paragraph{From staged falls to homes}
Surveys of fall detection~\cite{mubashir2013survey,igual2013challenges,noury2007fall,xu2018new,
wang2020elderly} agree that evaluation in real conditions is the weak point. Thirteen published
accelerometer algorithms caught 57\% of 29 real falls on average and raised 3--85 false alarms per day of
monitoring~\cite{bagala2012evaluation}; real falls of older people differ from those simulated by
students~\cite{klenk2011comparison}, and real fall recordings remain scarce~\cite{klenk2016farseeing}.
The few long in-home evaluations report false alarms in different units~\cite{broadley2018methods}: per
hour of use for a wearable alarm~\cite{kangas2015sensitivity}, and over years of continuous depth video
in 13 apartments~\cite{stone2015fall} and months of radar~\cite{liu2016automatic,su2018radar}; a camera
algorithm that
matched the state of the art on simulated data degraded sharply on months of home video~\cite{debard2016camera},
and an acoustic detector needed source height to bring false alarms down from 32 to 5 per
hour~\cite{popescu2008acoustic}. We therefore report false calls per day of ordinary living.

\paragraph{Multimodal LLMs as zero-shot judges of video}
Training-free video anomaly detection with language models started from captioning frames and scoring
the captions~\cite{zanella2024harnessing}; zero-shot multimodal LLMs tend to be conservative, with high
precision and collapsed recall, and class-specific instructions move the operating
point~\cite{yao2026multimodal}. For falls in visible-light video, open 8B models reach a fall F1 of about
0.68 in the wild but mostly call a fallen person ``lying'', and changing one prompt component moved fall
F1 by up to 13 points~\cite{schneider2025omnifall}; small VLMs prompted on RGB and depth stills are interpretable but
less accurate than specialised models~\cite{guendoul2025preliminary}; and Gemini Flash has been checked on
a small set of hospital fall clips~\cite{gabriel2026exposure}. On thermal imagery, benchmarks find that VLMs
fall back on language priors and degrade when the colour map changes~\cite{shrivastava2026thermeval,
moshtaghi2025rgbth,ashqar2024leveraging}, and six VLMs evaluated zero-shot on laboratory thermal sequences
of elder-care activities handled large movements but not safety-critical events such as
staggering~\cite{hu2026can}. To our knowledge, no prior study has evaluated a multimodal LLM as a zero-shot
fall screener on low-resolution thermal video with sound, varied the thermal resolution, rendering or
prompt for it, or reported its false-alarm burden over continuous recording in an occupied home.

\paragraph{Reliability of hosted models}
Hosted LLMs are not deterministic at temperature 0~\cite{song2025good,ouyang2025empirical}: across ten
identical runs of text tasks, accuracy varied by up to 15\%~\cite{atil2025nondeterminism}, which has been
traced to batch-size-dependent kernels in serving~\cite{he2025defeating}. Results also depend on prompt wording and
format~\cite{sclar2024quantifying,mizrahi2024state}. We measure both for the screening decision.

\section{The screening pipeline}
\label{sec:system}

\begin{figure}[t]
\centering
\includegraphics[width=\columnwidth]{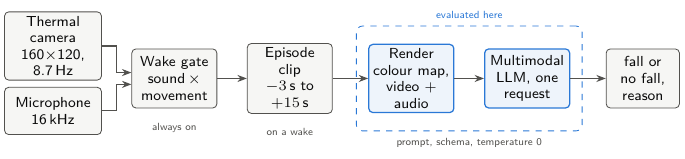}
\caption{The prototype. A thermal camera and a microphone stream continuously; a wake gate combines
sound and thermal movement; each wake becomes one short clip (3\,s before, 15\,s after, extended by
further wakes); the clip is rendered as a video with its sound track and reviewed once by a multimodal
LLM, which answers with a fixed JSON form. This paper evaluates the last two boxes.}
\label{fig:pipeline}
\end{figure}

We evaluate the decision stage of a research prototype of Senecta's in-home monitoring pod, which has run
continuously in one room since September 2026. \Cref{fig:pipeline} summarises it. The design rule is \emph{record always, analyse on
events}: sensors stream all the time, a cheap gate decides when something happened, and only then is a
short clip sent to the language model.

\subsection{Sensors and streaming}
A microcontroller (ESP32-S3) reads a radiometric FLIR Lepton~3.5 thermal camera~\cite{flir2023lepton}
($160\times120$ pixels, $\approx$\SI{57}{\degree} horizontal field of view, \SI{8.7}{\hertz}, absolute
temperature per pixel in centikelvin) and an I\textsuperscript{2}S MEMS microphone (INMP441,
\SI{16}{\kilo\hertz} mono). Frames are compressed losslessly on the device (median prediction with
Rice coding) and carry a CRC of the original pixels, so the server refuses rather than repairs a corrupted
frame. A server keeps the last minutes in memory and archives every 60\,s segment of raw packets to
object storage. The camera was a bench unit, not a finished product: for much of the recording it looked
through a thin polyethylene film window of unspecified composition, and it stood in two places, first
seeing most of the floor and later lower, behind a bed (\cref{sec:home}).

\subsection{Wake gate}
Every review costs money and time, so the gate decides what is worth a look. It combines a
\emph{sound score} (how far a microphone impulse rises above the room's adaptive noise floor, the 20th
percentile of recent levels; \SI{8}{\deci\bel} above it maps to 0 and \SI{24}{\deci\bel} to 1; sounds
without a sharp attack count half) with a \emph{movement score} (the changed fraction of a reduced
thermal frame, saturating at 10\%). The gate wakes when the product of the two reaches 0.15 with both
signals within \SI{2}{\second}, or when movement alone exceeds both 15\% of the frame and three times
the room's recent typical movement on two consecutive frames. A quiet, slow collapse while the room is
already busy does not wake it; we do not evaluate the gate's misses here.

\subsection{Episodes and clips}
A wake opens an episode covering \SI{3}{\second} before and \SI{15}{\second} after it, extended by
further wakes up to \SI{120}{\second}. Footage is reviewed in clips of at most \SI{30}{\second}; while
movement keeps an episode open, later clips are cut every \SI{20}{\second} of new footage or after
\SI{10}{\second} of quiet. A clip ends early at its last thermal frame when the camera stalls for more
than \SI{4}{\second} (the Lepton's flat-field correction holds the shutter for $\approx$\SI{1.8}{\second});
a clip whose audio coverage is below 85\% is reviewed without its sound track. A cap of 40 reviews an hour
holds back ordinary wakes in busy hours; strong wakes (a high index or large movement) are exempt.

\subsection{Rendering the thermal clip as video}
The model receives an MP4 rather than temperatures. Each clip is mapped to colour with one fixed scale
for the whole clip (the 2nd and 98th percentiles of all its pixels) through an 8-stop perceptually
ordered palette (black--purple--orange--pale yellow, ``inferno''), upscaled with nearest-neighbour
interpolation (a $160\times120$ frame becomes $480\times360$), and given a \SI{24}{px} strip with the
text \texttt{THERMAL t=}$t$\texttt{s}. Every source frame is kept at its own timestamp (variable frame
rate H.264, CRF 18) and the audio is high-passed at \SI{80}{\hertz}, de-clicked and AAC-encoded.

\subsection{The request}
\label{sec:request}
One request to Gemini~3.8~Flash (a model of the Gemini family~\cite{comanici2025gemini}; served by Vertex AI,
model identifier \texttt{gemini-3.8-flash}, October 2026) carries the video inline (sampled at 8 frames per
second), a fixed user text (``Review the complete $d$\,s thermal incident clip. Audio track present:
\ldots. Trigger time: \ldots. Only the footage is evidence.''), a three-line system instruction, and a
JSON schema; temperature is 0 and the model's thinking level is \textsc{low}. The deployed instruction,
called \emph{v8} below, is:
\begin{quote}\small\ttfamily
Review the thermal clip and audio.\\
Report fall\_observed only if you are very certain an upright person collapsed onto the floor.\\
A person who goes out of view, including dropping behind furniture or below the frame, is not a fall.
\end{quote}
The answer has three fields: \texttt{ends\_on} (floor, furniture, out\_of\_view, none, unknown), a
binary \texttt{verdict} and a free-text \texttt{reason}. One deterministic host rule applies: a fall
verdict whose person ends on furniture is turned into no fall (a heavy sit is not a fall). A fall verdict
starts a spoken check-in (``Are you okay?'') in the room; the check-in and the escalation that follows
are outside the scope of this paper.

The previous instruction, \emph{v5}, was deployed from 23 September until 4 October 2026 and is our main
comparison:
\begin{quote}\small\ttfamily
Review the thermal clip and audio.\\
Report fall\_observed only if an upright person visibly collapses onto the floor.\\
Normal motion, sitting, close-up blobs, and moving out of frame are no\_fall\_observed.
\end{quote}
Its schema asked for ten observation fields before the verdict (scene, person visible, before, after,
transition visible, audio observation, impact sound, visibility limits, evidence times, still supported)
and the client rejected a fall verdict that did not cite a visible transition. v8 replaced it after
replays of the bench recordings showed that v5 called most staged falls that ended out of the camera's
view a fall, and most deliberate lie-downs on the floor too. Neither prompt was tuned on the public dataset
used below; during development the deployed prompt was only checked on seven windows of that dataset's
development subjects.

\section{Data}
\label{sec:data}

\subsection{Staged falls in a laboratory: eHomeSeniors}
\label{sec:ehome}
The eHomeSeniors dataset~\cite{riquelme2019ehomeseniors} records six adults (three performers coached by
a physiotherapist, aged 34--37, and three adults aged 27--29) staging 15 fall types, five repetitions
each, in one $6\times5$\,m laboratory, with a Melexis MLX90640 thermopile array~\cite{melexis2018mlx90640} ($32\times24$ pixels,
\SI{55}{\degree}$\times$\SI{37.5}{\degree}) on the wall at \SI{1.2}{\metre}. We use the 90 Melexis
recordings (\num{\EhomeFrames} frames, median \SI{\EhomeFpsMedian}{\hertz}); the dataset has no audio.
Recordings are labelled only by fall type, with no frame-level times, and each holds several
repetitions, so we located every fall ourselves. An automatic detector proposed collapse times from the
height of the largest warm region; every one of the 90 recordings was then inspected on a strip of frames
sampled every \SI{0.5}{\second} and each repetition's landing time recorded (the moment the body reaches
floor level). The detector agreed with the inspection on \EhomeDetectorAgree{} of 90 recordings; slow
knee-flexion falls, falls towards or away from the camera, and the bed and chair falls were the usual
disagreements. One recording was excluded because its logger froze for \SI{84}{\second}. Temporal
annotation was done by a single annotator before any model saw any window (\cref{sec:ethics}).

From these times we cut 9\,s windows around each fall (\SI{3}{\second} before the landing to
\SI{6}{\second} after, shortened so that no window reaches into another repetition) and three kinds of
non-fall window of 5--9\,s that never contain a descent: \emph{upright} (standing, walking, turning,
before a fall or after getting up), \emph{recovery} (lying on the floor and getting up, a hard
negative for a screener that keys on ``person on the floor''), and \emph{rest} (lying on the bed before a
bed fall, sitting before a chair fall). This gives \EhomeFallWindows{} fall windows and
\EhomeNegWindows{} non-fall windows (\EhomeUpright{} upright, \EhomeRecovery{} recovery, \EhomeRest{}
rest). We keep the subject partition frozen before this work: subjects G1-1 and G2-1 for development,
G1-2 and G2-2 for validation, G1-3 and G2-3 held out. The held-out subjects were never inspected
during prompt development; with no training involved, the LLM results on all six subjects are also
reported. Windows are exported in the same event format the pipeline produces, without audio, and
rendered by the same code (\cref{sec:system}).

\subsection{Continuous recording in an occupied home}
\label{sec:home}
The prototype records one lived-in room (a bedroom with a desk). We use the eight days from
\HomeFirstDay{} to \HomeLastDay{}, when it recorded almost continuously and all staged sessions took place;
later recordings were intermittent, and from late September the bench camera developed calibration and
power faults. The window holds \num{\HomeSegments} archived minutes (\HomeRecordedHours{}\,h), of which
\HomeCameraHours{}\,h had a working camera. The camera delivered a median of \HomeFpsMedian{} frames per second and paused for its
flat-field correction about \HomeFfcPerHour{} times an hour (a hole of $\approx$\SI{1.8}{\second} each); a
warm body was in view in \HomeOccupiedPct\% of frames. The room was used normally: desk work, conversations, people passing, sitting and
resting, and people handling the prototype. Besides that, the authors staged falls in it on four days,
and on four days tested the prototype's response to a fall: someone dropped to the floor and then
role-played the resident's answer to the spoken check-in (``Actually, call Marcos'', ``Yeah, I'm not
okay''). The camera stood in two places:
\begin{itemize}
\item \emph{Floor in view} (6 September). The camera saw most of the floor. That afternoon nine falls
were staged onto the floor in view; several ended at the left or bottom edge of the frame, and one was
followed by a simulated long lie of about 20 minutes.
\item \emph{Floor hidden} (from 8 September). The camera stood lower, behind the bed. The 17 falls
staged on 8 and 10 September all ended out of its view, behind the bed or below the bottom edge of the
frame, and deliberate slow lie-downs on the floor on 9 September were partly hidden too.
\end{itemize}
The recording therefore answers four questions: how often the pipeline asks the model (wakes and clips
per hour); how often the model calls a fall during ordinary living, where every call is a false alarm;
how many of the visible staged falls it calls; and what it does when a fall happens but its landing is
hidden.

\begin{figure*}[t]
\centering
\includegraphics[width=\textwidth]{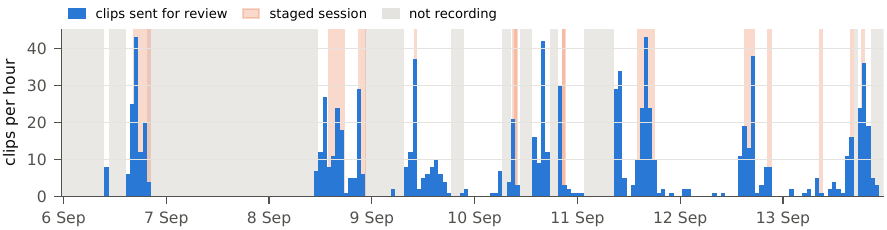}
\caption{The home recording, 6--13 September 2026: clips the deployed gate and pipeline sent for review, per
hour, with the staged sessions (staged falls, lie-downs and tests of the response) and the hours when the
bench unit was not recording.}
\label{fig:home}
\end{figure*}

We replayed the archive of 6--13 September (\cref{fig:home}) through the deployed gate and pipeline, in
recording time, with the
model call replaced by a recorder that answers ``no fall'' after \SI{8}{\second} (the median review
latency), and kept every clip the pipeline would have sent: \HomeClips{} clips, \HomeNormalClips{} of
them during ordinary living (\HomeNormalHours{}\,h of camera time outside staged sessions). Clips inside
staged sessions are labelled \emph{visible fall} when they show a staged fall the camera saw
(\HomeVisibleClips{} clips), \emph{aftermath} when they start with the person already on the floor after
one (\HomeAftermathClips{} clips), \emph{occluded fall} when they contain the moment of a staged fall the
camera did not see (\HomeOccludedClips{} clips) and \emph{lie-down} when they contain a deliberate
lie-down (\HomeLieClips{} clips); the remaining clips of staged sessions, three moments whose earlier labels
disagree and clips without a working camera are excluded. The 6 September session was not in the
project's records; we found it because the deployed model called four of its falls, and then labelled
every clip of that day on contact sheets without looking at the model's other answers. The later
sessions carry the labels recorded at the time by the people who staged them; of the 9 September
lie-downs we keep only the two whose labeller described them. The tests of the response were not in
those labels either; we found them in the prototype's own records of every check-in and contact call,
after a first analysis had counted them as ordinary living. Every check-in answered with a role-played
call for help, or a question about the fall, marks a test; answers less than 90 minutes apart form one
session, widened by 30 minutes on each side, and all of its clips are excluded (\HomeTestSessions{}
sessions, \HomeTestHours\,h). These sessions are defined from the answers alone, never from a model's
verdict, so excluding them cannot hide a false call; a check-in answered ``I'm fine'' marks nothing. One
more session was excluded on the authors' word: the afternoon of 11 September, when the response was
tested but the check-in could not be heard (its speaker was not ready), so no answer marks it. We asked
the authors about it because the deployed request's only fall call outside the staged sessions fell in
it, and we report that call (\cref{sec:results-home}); the afternoon holds \HomeConfirmedTestHours\,h of
camera time. Every fall call the language model made during ordinary living was checked on its frames
before being counted as a false alarm.

\section{Experimental protocol}
\label{sec:protocol}

\subsection{Conditions}
Every window and clip is reviewed under a baseline condition that reproduces the deployed request, and
under conditions that each change one factor (\cref{tab:conditions}). For the $160\times120$ home clips the
baseline is byte-identical to production: our harness produces the same MP4 (same SHA-256) and the same
request body as the deployed reviewer. For the $32\times24$ laboratory windows the baseline shows the frames
on the same $480\times360$ canvas as the home clips; the deployed code would have upscaled them only four
times, which is the \emph{display} condition.

\emph{Resolution} simulates a coarser sensor with the same field of view by averaging blocks of
radiometric pixels before colour mapping: in the home from $160\times120$ to $32\times24$ and
$8\times6$, and the $32\times24$ laboratory data to $16\times12$ and $8\times6$. It keeps the
temperature scale and adds no noise, so it is optimistic about real low-cost arrays, whose noise and
optics are worse; the $32\times24$ and $8\times6$ grids correspond to common thermopile arrays
(MLX90640~\cite{melexis2018mlx90640}, Grid-EYE AMG8833 at $8\times8$~\cite{panasonic2024gridye}). All
resolutions are displayed on the same
$480\times360$ canvas, so the model receives the same number of video pixels and tokens and only the
information differs. \emph{Display} instead keeps the deployed rule, which upscales by at most four
(a $32\times24$ array becomes a $128\times96$ video). \emph{Palette} replaces the colour map with
white-hot greyscale. \emph{Audio} removes the sound track (home clips only, for v8 and for v5). \emph{Prompt} either removes
the third sentence of v8 and nothing else (\emph{no out-of-view}) or uses the previous prompt, v5.
\emph{Model} runs the same request on two
earlier generations of the same family (Gemini~3.7 and 3.5 Flash). \emph{Repeat} sends the baseline
request again, unchanged, to measure test--retest stability at temperature 0. To see whether the main
effects hold beyond one model, the prompt, resolution and repeat conditions were also run on
Gemini~3.7 Flash.

During the study the Gemini~3.8 Flash endpoint slowed from about \SI{9}{\second} per request to
50--\SI{170}{\second}, often without an answer, while Gemini~3.7 Flash kept answering in 3--\SI{10}{\second}.
We therefore ran the full set of conditions on 3.7 Flash and the conditions that matter most on 3.8 Flash,
and ran the less important conditions on subsets fixed in advance: the two held-out subjects in the
laboratory, and in the home every clip of the staged sessions plus a fixed 20\% sample of the
ordinary-living clips, chosen by a hash of the clip identifier. \Cref{tab:conditions} lists what each
condition was run on.

\begin{table*}[t]
\centering
\caption{Conditions, each changing one factor of the deployed request, and what each was run on: laboratory windows
(all 1{,}001, or the 360 of the two held-out subjects) and home clips (the 395 ordinary-living clips, or
the fixed sample of 73 of them, together with the 48 staged clips; or the staged clips alone). ``--'': not run, or not
finished when the paper was built. Gemini~3.5 Flash was run with the base request on 360 laboratory windows.}
\label{tab:conditions}
\small
\begin{tabular}{@{}llcccc@{}}
\toprule
& & \multicolumn{2}{c}{Gemini 3.8 Flash} & \multicolumn{2}{c}{Gemini 3.7 Flash} \\
\cmidrule(lr){3-4}\cmidrule(l){5-6}
Condition & Change from the deployed request & lab & home & lab & home \\
\midrule
base & none: v8 prompt, inferno, $480\times360$, sound, \textsc{low} thinking & 1{,}001 & 443 & 1{,}001 & 121 \\
no out-of-view & v8 without its third sentence & 360 & 121 & 1{,}001 & 121 \\
v5 & previous prompt and answer form & 360 & 121 & 360 & 121 \\
repeat & identical request sent again & 1{,}001 & 121 & 360 & -- \\
res $32\times24$ & frames block-averaged to $32\times24$ (home) & -- & -- & -- & 121 \\
res $16\times12$ & frames block-averaged to $16\times12$ & 360 & -- & 360 & -- \\
res $8\times6$ & frames block-averaged to $8\times6$ & 1{,}001 & 121 & 360 & 121 \\
greyscale & white-hot greyscale instead of inferno & -- & -- & 360 & -- \\
display & deployed upscale rule ($\le 4\times$) & -- & -- & 360 & -- \\
no sound & sound track removed (home) & -- & 48 & -- & 48 \\
v5, no sound & v5, sound track removed (home) & -- & 48 & -- & 48 \\
\bottomrule
\end{tabular}
\end{table*}

The laboratory windows carry no sound track, and the request says so; their trigger time is set to the
window's start for every window, fall or not, so that it carries no information about the label.

\subsection{Outcomes}
On the laboratory windows we report \emph{sensitivity} (fall windows called a fall) and the
\emph{false-alarm rate} on each kind of non-fall window, with Wilson 95\% intervals~\cite{wilson1927}.
Windows from one recording share a subject and a fall type, so intervals that treat them as independent
are too narrow; we also report results per subject. On the home clips we report the number of fall
calls during ordinary living, per clip and per 24\,h of camera time, with exact Poisson 95\%
intervals~\cite{garwood1936}, and the share of visible and occluded staged falls called a fall. Paired
differences between a condition and the baseline use the exact McNemar test~\cite{mcnemar1947} on the same
items, and test--retest agreement Cohen's $\kappa$~\cite{cohen1960}. For the laboratory sensitivity we
also give an interval that resamples whole recordings (a cluster bootstrap over recordings), because the
repetitions in a recording share a subject and a fall type. A request that got no valid answer
was retried, up to six attempts and then in a later pass, until every case had an answer; an answer that
did not follow the response schema would count as no fall (there were \TotalInvalid{} such answers).

\subsection{Non-LLM baselines}
To put the zero-shot numbers in context we use two conventional detectors on the same windows, both
operating at $32\times24$ (home clips are block-averaged). A \emph{descent rule} flags a window when the
top of the largest warm region (pixels at least \SI{1.5}{\kelvin} above the frame median) drops by at
least a fraction $\theta$ of the image height within \SI{1.5}{\second} and stays down; $\theta$ is
chosen on the development subjects. A \emph{gradient-boosted classifier} uses 23 summary features of
the warm region's geometry over the window (drop, aspect ratio, height, width, centroid, area, excess
temperature, at the start and end of the window) and is trained on the development and validation
subjects. Both are then applied, unchanged, to the held-out subjects and to the home clips.

\section{Results}
\label{sec:results}

\subsection{Staged falls in the laboratory}
\label{sec:results-lab}
With the deployed request, Gemini~3.8 Flash called \SensBaseHeldout\% of the falls of the two held-out
subjects (\SensCountBaseHeldout) and \SensBaseAll\% over all six subjects (\SensCountBaseAll; 95\%
interval \SensClusterBaseAll\% when whole recordings are resampled), leaving out the bed falls discussed
below (\cref{tab:ehome}). It called none of the \FAnBaseAll{} non-fall windows a fall: not the
\EhomeCoreUpright{} of people walking, turning and sitting down, not the \EhomeCoreRecovery{} of a person
lying on the floor and getting up, and not the \EhomeCoreRest{} of a person lying still. Subjects differed (per-subject sensitivity \SubjectSensMin--\SubjectSensMax\%).
Bed falls were never called (\BedBase): at $32\times24$, lying on the bed and lying on the floor below it
differ by a few rows of pixels, and the only false calls the model made in the laboratory
(\FAcountBaseFull{} in all) were windows of the same recordings in which a person lay down on the bed.
Gemini~3.7 Flash, given the identical request, called more falls, \PairModelSensB\% against
\PairModelSensA\% on the same \PairModelN{} windows (McNemar $p\PairModelP$), and also no non-fall window.

\begin{figure}[t]
\centering
\includegraphics[width=\columnwidth]{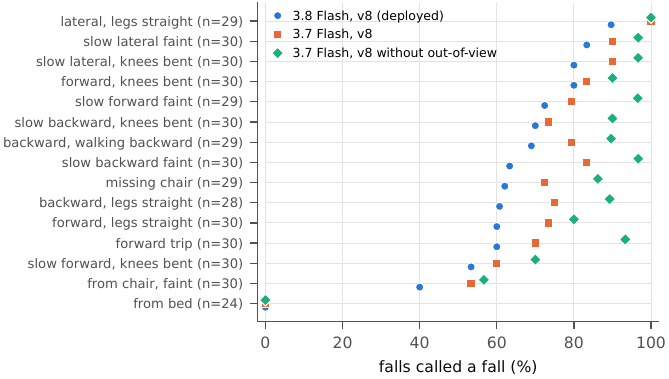}
\caption{Laboratory falls called a fall, by fall type (all six subjects): Gemini~3.8 Flash and 3.7 Flash
with the deployed prompt (v8), and 3.7 Flash with v8 without its out-of-view sentence. Bed falls are shown
but excluded from the other laboratory numbers.}
\label{fig:actions}
\end{figure}

\Cref{fig:actions} breaks sensitivity down by fall type. The model did best with falls that end in a
long, horizontal body in the middle of the frame (lateral falls), and worst with slow falls from a crouch
and with chair falls whose end posture resembles sitting.

\paragraph{Why falls were missed}
The model's reasons say why. Of the \MissTotal{} falls it missed (all subjects, bed falls excluded),
\MissLowering{} were described as a deliberate, controlled movement (``crouching'', ``kneeling'',
``lowering themselves'', ``sitting down''), and \MissOutOfView{} as the person going out of view below the
bottom edge of the frame, the situation the prompt's third sentence declares not to be a fall; in
\VetoLabDescribed{} of these the reason itself describes a fall (``collapses'', ``drops rapidly'') before
declining it. The staged protocol feeds the first group: performers get up within seconds, and the model
reads a fall followed by a quick recovery as a controlled movement. When the person was still on the floor
at the end of the window the model called \StillDownSens\% of falls (\StillDownCount), when they were
already back up \GotUpSens\% (\GotUpCount).

\paragraph{The model says when}
When the model called a fall, its reason almost always located it: in \GroundingBase\% of calls
(\GroundingCount) the first time or interval it named contained the annotated landing within
\SI{1}{\second}, so a person can check the answer by opening the clip at that moment.

\subsection{The occupied home}
\label{sec:results-home}
The gate woke \WakesTotal{} times, \WakesPerHour{} per camera-hour (\WakesSoundPct\% on sound with
movement, the rest on large movement alone). \WakesOpened{} wakes opened an episode; most of the others
extended one that was already open, and the pipeline's hourly cap on reviews held back \WakesBudget. The
pipeline produced \ClipsPerHour{} clips per camera-hour (\cref{fig:home}). During \HomeNormalHours{}\,h of
ordinary living, Gemini~3.8 Flash called \HomeBaseCallsWord{} of its \HomeBaseClips{} clips a fall (95\%
upper bound \HomeBasePerDayHi{} per 24\,h), and Gemini~3.7 Flash \HomeSevBaseCallsWord{} of the
\HomeSevBaseClips{} sampled clips (\cref{tab:home}). The excluded test afternoon of 11 September
(\cref{sec:home}) holds the only other clip the deployed request called a fall outside the staged
sessions (\HomeBaseConfirmedTestCalls{} of \HomeBaseConfirmedTestClips): a person walks to the back of the
room, drops behind the bed and stays out of view for about ten seconds, then comes back on all fours
towards the camera and stands up. The model answered that ``an upright individual suddenly collapses onto
the floor at around 10.4s accompanied by an impact sound'', while also answering that the person ended out
of view, which its prompt says is not a fall. Counted as ordinary living, that afternoon would make it one
call in \HomeNormalHoursWithConfirmed\,h.
The deployed request called
\HomeBaseVisible{} of the visible staged falls (3.7 Flash: \HomeSevBaseVisible), none of the clips that
began with a person already on the floor after a fall (\HomeBaseAftermath), none of the occluded staged
falls (\HomeBaseOccluded) and none of the deliberate lie-downs (\HomeBaseLie). The gate produced a clip
for \GateOccluded{} of the occluded staged falls and \GateLie{} of the lie-downs.

The missed visible falls are instructive. All but one ended at the edge of the frame, the body lying
along the left or bottom border, and the model's reason in each case invoked the prompt's third sentence
(\cref{tab:examples}): ``the person drops below the camera frame and goes out of view, which according to
the guidelines does not qualify as an observable fall''. Across visible and occluded home falls, the model
declined \VetoHomeOutOfView{} of its \VetoHomeMissed{} misses as out of view, and in
\VetoHomeDescribed{} of them its own reason describes the person losing balance or collapsing.

\begin{table*}[t]
\centering
\caption{The model's answers, verbatim (deployed request; the laboratory answers' opening ``Audio is unavailable'' is
omitted). Chosen to illustrate each situation, not sampled.}
\label{tab:examples}
\small
\begin{tabular}{@{}p{0.2\textwidth}p{0.11\textwidth}p{0.63\textwidth}@{}}
\toprule
Situation & Answer & Reason given by the model \\
\midrule
Laboratory fall, called (backward) & fall, \texttt{floor} & ``An upright person standing on the right side of the frame begins to collapse around t=2.6s, falling down to the floor where they remain lying horizontally in view through the end of the clip.'' \\
\addlinespace
Laboratory fall, missed (backward when trying to sit down) & no fall, \texttt{none} & ``An upright individual performs a controlled squat or crouch down to floor level around t=2.5s to t=3.5s, remains briefly in a low position, and then stands back up by the end of the clip without an uncontrolled collapse or fall.'' \\
\addlinespace
Laboratory fall, missed (backward) & no fall, \texttt{out\_of\_view} & ``The thermal footage shows a person standing in the center of the frame who descends downwards out of view below the bottom edge between 00:02.375 and 00:03.250. Because the subject moves out of view and does not remain clearly visible collapsing onto the floor within the frame, this cannot be confirmed as a fall.'' \\
\addlinespace
Home, visible fall, called & fall, \texttt{floor} & ``An upright individual walks across the room, loses balance around 22.8 seconds, and falls directly onto the floor with an audible impact at approximately 24.0 seconds, remaining on the ground.'' \\
\addlinespace
Home, visible fall, missed & no fall, \texttt{out\_of\_view} & ``The person drops below the camera frame and goes out of view, which according to the guidelines does not qualify as an observable fall.'' \\
\addlinespace
Home, occluded fall & no fall, \texttt{out\_of\_view} & ``The individual loses balance around 7.5 seconds and drops behind furniture out of view, which does not qualify as an observed fall onto the floor per guidelines.'' \\
\bottomrule
\end{tabular}
\end{table*}

\subsection{One sentence of the prompt}
\label{sec:results-prompt}
Removing only the third sentence of the deployed prompt (``A person who goes out of view, including
dropping behind furniture or below the frame, is not a fall'') raised laboratory sensitivity on both
models. On the held-out subjects, Gemini~3.8 Flash called \PairOovEightHoSensB\% of falls instead of
\PairOovEightHoSensA\% (\PairOovEightHoN{} falls, $p\PairOovEightHoP$) and \PairOovEightHoFaB{} of
\PairOovEightHoNegN{} non-fall windows instead of \PairOovEightHoFaA; Gemini~3.7 Flash \PairOovSevenHoSensB\%
instead of \PairOovSevenHoSensA\%, and over all six subjects \PairOovSevenSensB\% instead of
\PairOovSevenSensA\% (\PairOovSevenN{} falls, $p\PairOovSevenP$), with \PairOovSevenFaB{} of
\PairOovSevenNegN{} non-fall windows called a fall instead of \PairOovSevenFaA. The number of falls declined
as out of view dropped from \PairOovSevenOovA{} to \PairOovSevenOovB{} (3.7 Flash, all subjects). In the
home, without the sentence 3.8 Flash called \HomeNoOovVisible{} visible falls and \HomeNoOovOccluded{}
occluded falls, and \HomeNoOovCallsWord{} of the \HomeNoOovClips{} sampled ordinary-living clips, where the
deployed prompt called \HomeBaseSampleCalls; 3.7 Flash \HomeSevNoOovVisible, \HomeSevNoOovOccluded{} and
\HomeSevNoOovCalls{} (deployed prompt: \HomeSevBaseVisible, \HomeSevBaseOccluded{} and \HomeSevBaseCalls;
the one false call is the camera being moved). What removing the sentence cost in the home was the
deliberate lie-downs: 3.8 Flash called \HomeNoOovLie{} of them a fall (3.7 Flash \HomeSevNoOovLie).

The previous prompt, v5, which asked for ten observations before its verdict and named ``moving out of
frame'' among the non-falls, called \PairVfiveEightHoSensB\% of the held-out laboratory falls where v8
called \PairVfiveEightHoSensA\% (Gemini~3.8 Flash, \PairVfiveEightHoN{} falls, $p\PairVfiveEightHoP$), with
\PairVfiveEightHoFaB{} of \PairVfiveEightHoNegN{} non-falls called; on 3.7 Flash \PairVfiveSevenHoSensB\%
against \PairVfiveSevenHoSensA\%, with \PairVfiveSevenHoFaB{} non-falls called. In the home it called
\HomeVfiveVisible{} visible and \HomeVfiveOccluded{} occluded falls, and \HomeVfiveLie{} of the deliberate
lie-downs, and made \HomeVfiveCalls{} false calls on the sampled clips (3.7 Flash:
\HomeSevVfiveVisible, \HomeSevVfiveOccluded, \HomeSevVfiveLie{} and \HomeSevVfiveCalls). Calling the
lie-downs and the falls the camera did not see is what the out-of-view sentence was written to stop.
The sound mattered. Without the sound track, v5 called \HomeVfiveNoAudioOccluded{} occluded falls instead
of \HomeVfiveOccluded{} and \HomeVfiveNoAudioLie{} lie-downs instead of \HomeVfiveLie{} (3.7 Flash:
\HomeSevVfiveNoAudioOccluded{} instead of \HomeSevVfiveOccluded): where the landing could not be seen,
the sound of it was much of the evidence. v8 caught \HomeNoAudioVisible{} visible falls without sound
instead of \HomeBaseVisible{} (3.7 Flash: \HomeSevNoAudioVisible{} instead of \HomeSevBaseVisible).

\subsection{Resolution, rendering and model}
\label{sec:results-ablation}
\begin{figure}[t]
\centering
\includegraphics[width=\columnwidth]{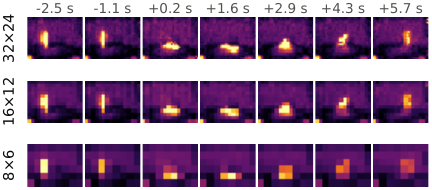}
\caption{One staged fall (a slow backward faint) from the public laboratory
data~\cite{riquelme2019ehomeseniors} (CC~BY~4.0), in the colour map the model saw, seconds from the
landing, at the sensor's $32\times24$ and block-averaged to $16\times12$ and $8\times6$. The performer is
back on their feet within six seconds.}
\label{fig:examples}
\end{figure}

\begin{figure}[t]
\centering
\includegraphics[width=\columnwidth]{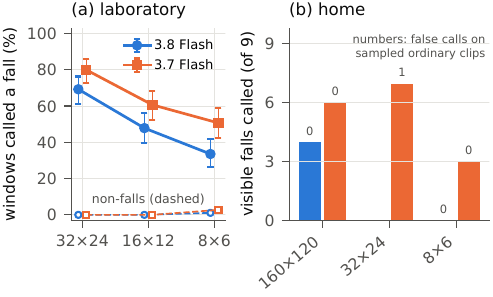}
\caption{Effect of sensor resolution (block averaging, same field of view, same display size).
(a) Laboratory falls and non-falls called a fall, held-out subjects, Wilson 95\% intervals.
(b) Home: visible staged falls called (of 9) and false calls on sampled ordinary-living clips.}
\label{fig:resolution}
\end{figure}

\Cref{fig:examples} shows what the coarser grids leave of a fall, and \cref{fig:resolution} the effect on
the verdicts. In the laboratory, on the held-out subjects, the deployed 3.8 Flash called
\PairResSixteenEightHoSensB\% of falls at $16\times12$ and \PairResEightEightHoSensB\% at $8\times6$, the
size of the cheapest arrays, against \PairResEightEightHoSensA\% at the sensor's $32\times24$; Gemini~3.7
Flash \PairResSixteenSevenHoSensB\% and \PairResEightSevenHoSensB\% against \PairResSixteenSevenHoSensA\%.
At $8\times6$ the models called \PairResEightEightHoFaB{} (3.8 Flash) and \PairResEightSevenHoFaB{}
(3.7 Flash) of the \PairResEightEightHoNegN{} non-fall windows a fall, against none at $32\times24$. The
coarse grids changed the model's description more than its caution: over all subjects, 3.8 Flash declined
\PairResEightEightOovB{} falls as out of view at $8\times6$, against \PairResEightEightOovA{} at full
resolution, because a person reduced to a few warm blocks near the bottom of the frame looks like someone
leaving it. In the home (3.7 Flash, sampled clips), false calls were \HomeSevBaseCalls{} at
$160\times120$, \HomeSevResThirtyTwoByTwentyFourCalls{} at $32\times24$ and \HomeSevResEightBySixCalls{}
at $8\times6$, and the visible falls called went from \HomeSevBaseVisible{} to
\HomeSevResThirtyTwoByTwentyFourVisible{} and \HomeSevResEightBySixVisible. The deployed model at
$8\times6$ caught \HomeResEightBySixVisible{} of the visible falls, with \HomeResEightBySixCalls{} false
calls on the sampled clips.

The rendering mattered less. On Gemini~3.7 Flash and the held-out subjects, a greyscale colour map
instead of inferno gave \PairGraySevenHoSensB\% laboratory sensitivity against \PairGraySevenHoSensA\%,
and the deployed display rule, which shows a $32\times24$ array as a $128\times96$ video instead of
$480\times360$, \SensSevDisplayHeldout\% against the same \PairGraySevenHoSensA\%. The older Gemini~3.5 Flash, given the deployed request, called
\SensFlashThreeFiveHeldout\% of the held-out falls, more than 3.7 Flash (\SensSevBaseHeldout\%) and 3.8
Flash (\SensBaseHeldout\%), and \FAFlashThreeFiveHeldout\% of the non-fall windows (3.8 Flash:
\FABaseHeldout\%).

\subsection{The same question, asked again}
\label{sec:results-retest}
Temperature 0 did not make the answer reproducible (\cref{tab:retest}). Two identical laboratory runs of
the deployed request agreed on \RetestLabAgree\% of \RetestLabN{} windows (Cohen's
$\kappa=\RetestLabKappa$; Gemini~3.7 Flash, held-out subjects: \RetestSevLabAgree\% of \RetestSevLabN,
$\kappa=\RetestSevLabKappa$); the \RetestLabFlips{} windows whose verdict changed were almost all falls,
the borderline cases a single request cannot settle, and overall sensitivity barely moved because flips
went both ways. This can be used: calling a fall when either of two identical requests does raised
laboratory sensitivity from \EitherLabFirstSens\% to \EitherLabSens\% with \EitherLabFA{} of \EitherLabN{}
non-fall windows called. In the home it called \EitherHomeCalls{} of the \EitherHomeClips{} sampled
ordinary-living clips a fall (one request: \EitherHomeFirstCalls) and \EitherHomeVisible{} of the visible
falls (one request: \EitherHomeFirstVisible).

\subsection{Conventional detectors}
\label{sec:results-baselines}
\begin{figure}[t]
\centering
\includegraphics[width=\columnwidth]{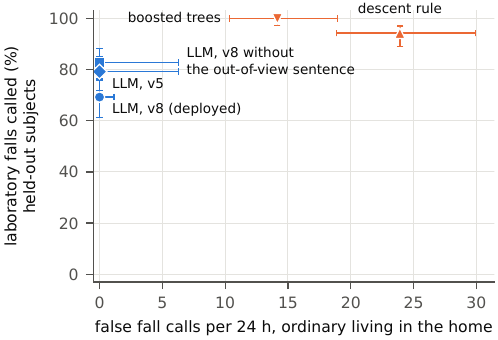}
\caption{Each screener as one point: laboratory falls called (held-out subjects, Wilson 95\% interval)
against false fall calls per 24\,h of ordinary living in the home (exact Poisson 95\% interval). The
conventional detectors were built on the laboratory data and run at $32\times24$.}
\label{fig:tradeoff}
\end{figure}

\Cref{tab:compare} and \cref{fig:tradeoff} set the model beside two detectors built on the laboratory
data. On the held-out subjects the boosted trees called \SensTreesHeldout\% of falls and the descent rule
\SensRuleHeldout\%, with \FATreesHeldout\% and \FARuleHeldout\% of non-fall windows called, far ahead of the
zero-shot model's \SensBaseHeldout\% at \FABaseHeldout\%. Taken to the home without change, they caught as
many or more of the visible falls (\HomeTreesVisible{} and \HomeRuleVisible) but called a fall on
\HomeTreesCalls{} and \HomeRuleCalls{} of the \HomeTreesClips{} ordinary-living clips, \HomeTreesPerDay{}
and \HomeRulePerDay{} false alarms per 24\,h, and they also called many of the clips that began with a
person already on the floor.

\begin{table*}[t]
\centering
\caption{Laboratory windows (eHomeSeniors, $32\times24$, no sound), 14 fall types (bed falls apart, \cref{sec:results-lab}).
Falls called (\%) for the two held-out subjects ($n=140$) and for all six subjects ($n=414$; ``--'' where a condition was
run on the held-out subjects only), and non-fall windows called a fall (upright, recovery and rest windows together; all
subjects, or held-out where only those were run).}
\label{tab:ehome}
\small
\begin{tabular}{@{}lcccccc@{}}
\toprule
& \multicolumn{3}{c}{Gemini 3.8 Flash (deployed)} & \multicolumn{3}{c}{Gemini 3.7 Flash} \\
\cmidrule(lr){2-4}\cmidrule(l){5-7}
Condition & held-out & all & non-falls & held-out & all & non-falls \\
\midrule
v8, deployed & 69 & 67 & 0/499 & 80 & 77 & 0/499 \\
v8 without the out-of-view sentence & 83 & -- & 0/182 & 92 & 88 & 4/499 \\
v5 (previous prompt) & 79 & -- & 2/182 & 81 & -- & 0/182 \\
same request, sent again & 72 & 68 & 0/499 & 76 & -- & 1/182 \\
resolution $16\times12$ & 48 & -- & 0/182 & 61 & -- & 0/182 \\
resolution $8\times6$ & 34 & 29 & 5/499 & 51 & -- & 5/182 \\
greyscale colour map & -- & -- & -- & 73 & -- & 1/182 \\
deployed display, $128\times96$ & -- & -- & -- & 86 & -- & 0/182 \\
\bottomrule
\end{tabular}
\end{table*}

\begin{table*}[t]
\centering
\caption{The occupied home ($160\times120$ with sound). False calls: ordinary-living clips called a fall, all of them
false alarms, out of the clips reviewed (every clip for the deployed request on Gemini~3.8 Flash; a fixed 20\% sample
otherwise; ``--'' for the conditions run on the staged clips only).
Visible: staged falls the camera saw, called. Occluded: staged falls whose landing the camera did not see, called.
Clips that began with a person already on the floor, and deliberate lie-downs, are reported in the text.}
\label{tab:home}
\small
\begin{tabular}{@{}lcccccc@{}}
\toprule
& \multicolumn{3}{c}{Gemini 3.8 Flash (deployed)} & \multicolumn{3}{c}{Gemini 3.7 Flash} \\
\cmidrule(lr){2-4}\cmidrule(l){5-7}
Condition & false calls & visible & occluded & false calls & visible & occluded \\
\midrule
v8, deployed & 0/395 & 4/9 & 0/15 & 0/73 & 6/9 & 2/15 \\
v8 without the out-of-view sentence & 0/73 & 9/9 & 13/15 & 1/73 & 9/9 & 11/15 \\
v5 (previous prompt) & 0/73 & 5/9 & 10/15 & 0/73 & 6/9 & 11/15 \\
v8, sound removed & -- & 2/9 & 0/15 & -- & 5/9 & 0/15 \\
v5, sound removed & -- & 6/9 & 5/15 & -- & 8/9 & 8/15 \\
same request, sent again & 0/73 & 3/9 & 0/15 & -- & -- & -- \\
resolution $32\times24$ & -- & -- & -- & 1/73 & 7/9 & 3/15 \\
resolution $8\times6$ & 0/73 & 0/9 & 1/15 & 0/73 & 3/9 & 1/15 \\
\bottomrule
\end{tabular}
\end{table*}

\begin{table*}[t]
\centering
\caption{The zero-shot model and two conventional detectors. Laboratory: held-out subjects, 14 fall types, falls and
non-falls called (\%). Home: visible staged falls called, ordinary-living clips called a fall (all false alarms), and
occluded staged falls called. The detectors run at $32\times24$ and were built on the laboratory data.}
\label{tab:compare}
\small
\begin{tabular}{@{}lccccc@{}}
\toprule
& \multicolumn{2}{c}{Laboratory (held-out), \%} & \multicolumn{3}{c}{Home, clips called a fall} \\
\cmidrule(lr){2-3}\cmidrule(l){4-6}
Screener & falls & non-falls & visible falls & ordinary living & occluded falls \\
\midrule
3.8 Flash, v8 (deployed) & 69 & 0.0 & 4/9 & 0/395 & 0/15 \\
3.8 Flash, no out-of-view & 83 & 0.0 & 9/9 & 0/73 & 13/15 \\
3.7 Flash, v8 & 80 & 0.0 & 6/9 & 0/73 & 2/15 \\
3.7 Flash, no out-of-view & 92 & 1.1 & 9/9 & 1/73 & 11/15 \\
descent rule & 94 & 1.1 & 7/9 & 76/395 & 13/15 \\
boosted trees (trained) & 100 & 1.1 & 6/9 & 45/395 & 8/15 \\
\bottomrule
\end{tabular}
\end{table*}

\begin{table}[t]
\centering
\caption{Test--retest at temperature 0: the deployed request sent twice to the same model. Agreement (\%),
Cohen's $\kappa$ and the number of windows or clips whose verdict flipped.}
\label{tab:retest}
\small
\begin{tabular}{@{}llcccc@{}}
\toprule
Data & Model & $n$ & agree & $\kappa$ & flips \\
\midrule
laboratory & 3.8 Flash & 1,001 & 95.9 & 0.90 & 41 \\
laboratory & 3.7 Flash & 360 & 95.8 & 0.91 & 15 \\
home & 3.8 Flash & 121 & 97.5 & 0.56 & 3 \\
\bottomrule
\end{tabular}
\end{table}

\subsection{Latency and cost}
\label{sec:results-cost}
A laboratory window (\SI{9}{\second}, no sound) cost \TokensLab{} input tokens and a home clip
(\SI{18}{\second} with sound) \TokensHome; at the list price this is \$\CostLab{} and \$\CostHome{} per
review. The median request to Gemini~3.8 Flash took \SI{\LatencyLab}{\second} (laboratory) and
\SI{\LatencyHome}{\second} (home), with a 90th percentile of \SI{\LatencyNinetyLab}{\second} and
\SI{\LatencyNinetyHome}{\second}. During part of the study the same endpoint took minutes per request or
did not answer at all (\cref{sec:protocol}), which is what a check-in that waits for the verdict would
have experienced. The study used \StudyTokens{} million input tokens, about \$\StudyCost{}.

\section{Discussion}
\label{sec:discussion}

\paragraph{The prompt is the operating point}
A conventional detector has a threshold; a screener built on a language model has prose. The third
sentence of the deployed prompt was written for one installation, a low camera behind a bed, where
staged falls disappeared from view and the previous prompt inferred them anyway from a descent and a
thud. In that room it did what it was meant to. Elsewhere it is a statement about geometry that the
model applies literally: a person lying along the bottom edge of a $32\times24$ frame, or at the left
edge of a camera that sees the floor, ``goes out of view'', and the model says so and declines. A
policy sentence in a prompt therefore needs testing on every camera placement it will meet, like a
threshold on every dataset. Here it cost laboratory sensitivity (\PairOovEightHoSensA\% with it,
\PairOovEightHoSensB\% without) and visible falls at home (\HomeBaseVisibleK{} of \HomeVisibleClips{}
with it, \HomeNoOovVisibleK{} without), and what it bought was narrow: in ordinary living 3.8 Flash
made no false call with or without it, so its only gain in the home was not calling the deliberate
lie-downs. A sturdier design keeps policy out of the prompt: ask the model for
observations it can be checked on (where the person ends, whether the landing was visible, when it
happened), and apply the policy in code, where it can depend on the known geometry of the installation.
The deployed form already does this for one case, a fall onto furniture.

\paragraph{Zero-shot versus trained}
In the laboratory the trained classifier wins by a wide margin. It has seen these six people, this room
and this sensor, and the zero-shot model has seen none of them. In the home the order of the two numbers
that matter reverses. The trained detectors still catch more of the visible falls, but they called a
fall \HomeTreesPerDay{} and \HomeRulePerDay{} times per 24\,h of ordinary life, which no household would
tolerate; the zero-shot model made \HomeBaseCallsWord{} in \HomeNormalHours\,h, without any data from
that room. The two are complementary, and the prototype's
architecture already reflects it: a sensitive, cheap gate decides that something happened, and the model
vetoes most of what is not a fall. What this study adds is a measurement of the price of that veto in
missed falls, and of how much of that price is set by a single sentence.

\paragraph{How many pixels}
Every halving of resolution below the laboratory sensor's $32\times24$ cost falls, by about the same
amount on both models. On the deployed model the held-out falls called went from
\PairResEightEightHoSensA\% to \PairResSixteenEightHoSensB\% at $16\times12$ and
\PairResEightEightHoSensB\% at $8\times6$, where a person is a few warm blocks; on 3.7 Flash from
\PairResSixteenSevenHoSensA\% to \PairResSixteenSevenHoSensB\% and \PairResEightSevenHoSensB\%. Low resolution made the model invent few falls: at $8\times6$ it called
\PairResEightEightHoFaB{} (3.8 Flash) and \PairResEightSevenHoFaB{} (3.7 Flash) of
\PairResEightEightHoNegN{} non-fall windows a fall, against none at full resolution. Mostly it lost
falls instead, because a coarse figure near the edge of the frame reads as someone leaving it, which the
deployed prompt rules out as a fall. Cheaper sensors therefore cost mostly sensitivity. For
hardware, arrays of $32\times24$ and finer are usable for this kind of review and $8\times8$ arrays are
not, at least with a general-purpose model and no training; between the two, a sensor buys sensitivity
with every pixel.

\paragraph{Model versions move the operating point}
The same request, sent to three generations of one model family, called \SensFlashThreeFiveHeldout\%,
\SensSevBaseHeldout\% and \SensBaseHeldout\% of the held-out falls (Gemini~3.5, 3.7 and 3.8 Flash), with
no laboratory false alarm from any of them: each newer model was more cautious. A deployment that pins
a model version keeps the operating point it measured; one that follows a provider's updates moves it
without notice, and needs these tests rerun with every update.

\paragraph{After the fall}
The model did not flag a person already lying on the floor when the clip did not show the collapse,
including a simulated long lie of about twenty minutes, because the prompt asks about a collapse. For the
people a screener is for, the long lie is the dangerous part~\cite{fleming2008inability}. A screener
should ask a second, separate question (is someone on the floor now, and for how long), rather than rely
on catching the moment of the fall.

\paragraph{Reproducibility}
At temperature 0 a few percent of verdicts change when the same request is sent again, concentrated on
the falls that are hardest to judge. Majority voting cannot use this, because flips are rare and go both
ways; calling a fall when either of two answers does recovers some of the borderline falls, at the price
of a second request and a little more risk of false calls. What a deployment needs in any case is the
record, the clip, the request and the answer, kept for every decision, and an answer that can be checked. The model's reasons locate the event in time almost every
time it calls a fall, so a person can verify a decision in seconds.

\subsection{Limitations}
\label{sec:limitations}

\emph{The falls are staged.} All falls were performed on purpose by adults much younger than the people a
screener is for: six performers in a laboratory and members of our team in the home. Real falls of older people are slower,
more varied and often happen at night, beside furniture, in bathrooms~\cite{bagala2012evaluation}. The
laboratory protocol also makes performers get up within seconds, which made several falls look
deliberate to the model.

\emph{One home, one camera at a time, one model family.} The home recording is one room over eight days;
the camera saw the floor on one day only, so the nine visible home falls give a wide interval. Sensitivity
in a home where the camera is mounted high and sees the whole floor is the most important number we could
not measure. All models are Gemini Flash generations served by one endpoint; models change without
notice, the same request can get a different answer, and the deployed model's endpoint slowed so much
during the study that several conditions could be run only on its predecessor
(\cref{tab:conditions}).

\emph{Labels.} The fall times of the public dataset were located by one annotator, AI-assisted, on
$0.5$\,s frame strips; bed falls could not be annotated reliably and are reported apart. The home
labels of 8--10 September were made at the time by the people who staged them, the tests of the
response were found afterwards in the prototype's own records, and the 6 September session was labelled
by inspecting every clip. One test afternoon, 11 September, was excluded on the authors' word after a
model's call in it had been seen; counted as ordinary living, it adds that one call (\cref{sec:results-home}).
Unrecorded staged activity may remain among the ordinary-living clips; it can only make the false-call
counts higher than they should be, since every model call there is counted, or hide a miss.

\emph{Simulated resolution.} Block averaging a microbolometer image gives a lower-resolution image with
the noise and optics of a much better sensor. Real thermopile arrays are noisier and have their own
artefacts, so the low-resolution numbers are an upper bound for those devices.

\emph{The gate is not evaluated.} We review the clips the deployed gate produced. A fall that does not
wake it is never reviewed; its own misses need recordings with the floor in view and many more falls.

\section{Ethics, data and disclosure}
\label{sec:ethics}

\textbf{Home recordings.} The home recordings come from a research prototype installed by the authors
in a room they use, and everyone who used the room knew that it recorded heat and sound. The staged falls,
lie-downs and tests of the response were performed by members of the team, onto the floor of their own
room. The recordings are not released: they contain the voices
and private conversations of the people in the room. No image of the home appears in this paper.

\textbf{Public data.} eHomeSeniors~\cite{riquelme2019ehomeseniors} is published as supplementary
material under CC~BY~4.0. We used only its thermal frames; no participant metadata or file names were
sent to the model. Our changes are the cut windows, the conversion to centikelvin and the video
rendering. We release the frame-level fall times we annotated for its Melexis recordings, with the windows,
prompts and model answers (\url{https://senecta.ai/research/thermal-fall-llm-data.zip}).

\textbf{Cloud processing.} Every clip was sent to a commercial model through Vertex AI. Thermal video
at these resolutions does not show faces, but the home clips carry speech, and a deployed system would
send audio from people's homes to a third party; that requires the residents' informed consent and a
data-processing agreement, and in some jurisdictions the consent of everyone recorded.

\textbf{Scope of the claims.} The prototype is a research system, not a medical device. Nothing here
estimates how often real falls of older people would be recognised; the staged falls were performed by
younger adults, in one laboratory and one room.

\textbf{Competing interests.} Both authors are founders of Senecta Inc., which develops the in-home monitoring
service for home care agencies whose prototype is evaluated here (\url{https://senecta.ai}); Senecta funded
the work.

\textbf{Use of AI assistance.} The temporal annotation of the public dataset, the experiment and analysis
code and an initial draft of this manuscript were produced with the assistance of an AI model (Claude,
Anthropic) under the authors' direction. The authors take full responsibility for the content.

\section{Conclusion}
\label{sec:conclusion}

Used zero-shot as the decision stage of an event-triggered thermal screener, a production multimodal
LLM made almost no false fall calls in ordinary home life and in laboratory recordings, explained when
it did call a fall, and needed no data from the room it watched. It also missed a large share of clearly
staged falls, and how large was decided less by the sensor than by one cautionary sentence in its prompt,
written for a different camera placement. Each halving of resolution cost falls but added few false
calls; colour map and display size mattered less; asking again changed a few percent of answers. A
detector trained on laboratory falls was far more sensitive in that laboratory and raised false alarms
many times a day in a home. For anyone building on such a model, the practical advice is to measure
false alarms in real homes and sensitivity on falls the camera can see, to keep decision policy in code
rather than in the prompt, and to treat the prompt as a parameter that is evaluated, versioned and
re-tested whenever the installation or the model version changes.

\bibliographystyle{IEEEtran}
\bibliography{refs}

\end{document}